\documentclass{article}

\usepackage{PRIMEarxiv}

\usepackage[utf8]{inputenc} % allow utf-8 input
\usepackage[T1]{fontenc}    % use 8-bit T1 fonts
\usepackage{hyperref}       % hyperlinks
\usepackage{url}            % simple URL typesetting
\usepackage{booktabs}       % professional-quality tables
\usepackage{amsfonts}       % blackboard math symbols
\usepackage{nicefrac}       % compact symbols for 1/2, etc.
\usepackage{microtype}      % microtypography
\usepackage{lipsum}
\usepackage{fancyhdr}       % header
\usepackage{graphicx}       % graphics
\graphicspath{{media/}}     % organize your images and other figures under media/ folder
\usepackage{amsmath}

\title{Efficient Slice Anomaly Detection Network for 3D brain MRI Volume
}

\author{
  Zeduo Zhang \\
  Department of Computer Science \\
  Western University \\
  London, On, Canada\\
  Vector Institute for Artificial intelligence\\
  Toronto, On, Canada\\ 
  \texttt{zzhan762@uwo.ca} \\
   \And
  Yalda Mohsenzadeh \\
  Department of Computer Science \\
  Western University \\
  London, On, Canada\\
  Vector Institute for Artificial intelligence\\
  Toronto, On, Canada\\
  \texttt{ymohsenz@uwo.ca} \\
}

\begin{document}
\maketitle

\begin{abstract}
Current anomaly detection methods excel with benchmark industrial data but struggle with natural images and medical data due to varying definitions of 'normal' and 'abnormal.' This makes accurate identification of deviations in these fields particularly challenging.
Especially for 3D brain MRI data, all the state-of-the-art models are reconstruction-based with 3D convolutional neural networks which are memory-intensive, time-consuming and producing noisy outputs that require further post-processing.
We propose a framework called Simple Slice-based Network (SimpleSliceNet), which utilizes a model pre-trained on ImageNet and fine-tuned on a separate MRI dataset as a 2D slice feature extractor to reduce computational cost. We aggregate the extracted features to perform anomaly detection tasks on 3D brain MRI volumes. Our model integrates a conditional normalizing flow to calculate log likelihood of features and employs the Semi-Push-Pull Mechanism to enhance anomaly detection accuracy. 
The results indicate improved performance, showcasing our model's remarkable adaptability and effectiveness when addressing the challenges exists in brain MRI data. In addition, for the large-scale 3D brain volumes, our model SimpleSliceNet outperforms the state-of-the-art 2D and 3D models in terms of accuracy, memory usage and time consumption. Code is available at: https://anonymous.4open.science/r/SimpleSliceNet-8EA3.
\end{abstract}

% keywords can be removed

\section{Author summary}
In this work, we address the need for more effective methods to identify unusual patterns in brain MRI scans, which are crucial for diagnosing neurological conditions. Traditional techniques often struggle with the high diversity in normal brain appearances and are usually slow and demanding on computing resources. To overcome these challenges, we developed SimpleSliceNet, a streamlined approach that simplifies the analysis without sacrificing accuracy.
Our method operates more quickly and requires fewer computational resources than older methods, making it highly practical for clinical settings. By improving the detection of anomalies in brain images, our method holds promise for earlier and more accurate diagnoses of neurological issues, which can directly enhance patient care. Additionally, our approach reduces the technical barriers often faced in medical imaging, potentially broadening the access to high-quality diagnostic techniques across different healthcare environments.
Our findings demonstrate that our approach surpasses current leading methods, suggesting a significant step forward in the field of medical imaging.

\section{Introduction}
\label{sec:intro}
% background for UAD
Anomaly detection (AD) aims to identify defects or outliers within datasets, commonly applied across natural, industrial, and medical data. Unsupervised anomaly detection (UAD) \cite{zhang2023destseg,gudovskiy2022cflow,guo2023template,jewell2022one} has gained attraction due to the diverse and unpredictable nature of anomalies. In addition, the scarcity of anomaly samples poses challenges for model adaptation, often leading to overfitting. Thus, UAD, which requires no anomaly sample training, is often preferred. %Despite their scarcity, these anomaly samples contain valuable information.

% background for brain MRI dataset and state the problem
Magnetic Resonance Imaging (MRI) of the brain is a non-invasive medical imaging technique used to produce detailed images of brain structure and function. Anomalies in brain MRI indicate deviations from typical or healthy brain structure or function, including abnormalities like tumors, cysts, or vascular malformations. While the MVTecAD production line dataset \cite{bergmann2019mvtec} is commonly used to evaluate state-of-the-art anomaly detection models for images, it differs from brain MRI data. In the MVTecAD dataset, normal objects exhibit consistent patterns characterized by concentrated normal features, and any deviations from these patterns are identified as anomalies. However, brain MRI structure can vary due to differences among patients, biological changes, technical factors, patient movement, and calibration and correction processes.

% state the problem and review previous work
While the state-of-the-art methods \cite{liu2023simplenet,yao2023BGAD,roth2022towards,gudovskiy2022cflow,salehi2021multiresolution} have achieved nearly perfect outcomes on industrial data (e.g. MVTecAD), they still face challenges when applied on natural images and medical data. Prominent models, including reconstruction-based models \cite{guo2023EDC}, embedding-based models \cite{liu2023simplenet,yao2023BGAD,gudovskiy2022cflow}, and student-teacher models \cite{zhang2023destseg,deng2022reverseST,tien2023revisitingRST}, often employ feature extractors pre-trained on ImageNet. This dataset which includes natural object images features a domain distinct from medical data. Furthermore, models specifically designed for 3D data are scarce. The most common method to perform anomaly detection on 3D brain MRI volumes are reconstruction-based frameworks with 3D Convolutional Neural Network (3D-CNN) \cite{luo2023uad3dbrain,pinaya2022uad3dbrain_transformer}. These models require lots of memory to store the 3D kernels, consume time to converge, and typically produces noisy results, necessitating additional post-processing steps. Few supervised anomaly detection methods \cite{rudie2021UNETbrain,li2021UNETwhitematter} for 3D data have also been proposed, presenting the challenge of costly annotation for such data. Undoubtedly, they also encounter the same issue of memory and time.

% possible solution and link to contribution
To harness the capabilities of existing 2D anomaly detection architectures, we found that methods using 2D-slice models are more practical \cite{gupta2021slicemodel,gupta2023slicemodelpretrained}. These models process each slice in the volume sequentially and aggregate the information afterward. Considering the considerable volume of 3D data, certain complex and computationally demanding models may not be feasible. We seek a simple yet effective model suitable for both 2D and 3D brain MRI data due to the significant volume of 3D data and computational constraints. Recently presented SimpleNet \cite{liu2023simplenet} emerges as a promising candidate, being lightweight, efficient, and not requiring anomalies for training. However, its instability in certain scenarios and applicability only to single-structure images, like those in industry production lines, pose challenges. To adapt SimpleNet to the diverse features of brain MRI, particularly in distinguishing between normal and abnormal features, modifications are necessary.

% contribution
% why use simplenet
The primary objective of this study is to find an efficient model for detecting anomalies in brain MRI images and apply it to analyze 3D brain MRI volumes. We propose a novel anomaly detection model for medical data inspired by SimpleNet framework \cite{liu2023simplenet} and evaluate the effectiveness and performance of our model on both 2D and 3D brain MRI datasets for this purpose. 
% why use pre-trained and EDC
To ensure adaptability to medical data, we fine-tuned the feature extractor (which was pre-trained on ImageNet) on an independent medical dataset with an unsupervised Encoder-Decoder Contrast (EDC) \cite{guo2023EDC} technique. This step also reduces computational cost during training of our SimpleSliceNet model.
% why use BG-SPP loss
Additionally, we propose that the discriminator in SimpleNet \cite{liu2023simplenet}, which utilizes truncated L1 loss, functions as a complete pushing mechanism that could potentially contribute to a higher false positive rate. To tackle this issue, we instead introduce a Semi-Push-Pull contrastive Loss to our model, which selectively modifies ambiguous regions to establish an ambiguous boundary between normal and abnormal features, thereby reducing false positives. While anomalies in SimpleNet are synthesized by adding noise to the feature, categorizing those with little variation solely as anomalies may lead to a high false positive rate. We discover that the Semi-Push-Pull Mechanism proves beneficial in addressing this issue by selectively modifying the ambiguous region. So that the anomalous features won't be forced to deviate from the normal distribution too much. Instead, an ambiguous boundary between the normal and abnormal features is established. And so those synthesized anomalies that are close to anomalies still have high likelihood. 
% why use CNF
Furthermore, we found that the training process in SimpleNet is unstable which might be caused by the unclear balance between projection and discriminator layers and the full pushing mechanism. We show that a single Conditional Normalizing Flow (CNF) can address this issue and outperform without additional projection layers.
% why use 2D-slice instead of 3D CNN
Finally, we demonstrated the effectiveness of our model in detecting anomalies in 3D volumes using a slice-by-slice approach, leveraging the idea of 2D-Slice models. Few papers focus on applying embedding-based methods to medical datasets due to the aforementioned challenges and issues presented in \nameref{sec:embed}, especially for 3D brain MRI volumes, where conventional 2D models cannot be directly applied. The contributions of this paper are summarized as below: 
\begin{enumerate}
    \item We design new lightweight embedding-based model with a pre-trained features extractor and an additional semi-push-pull loss to detect and localize the anomalies on brain MRI dataset and outperform the state-of-the-art (SOTA) models.
    \item We highlight the effectiveness of fine-tuning the feature extractor model using a separate MRI dataset, a choice made to ensure the pre-trained model remains uninformed about knowledge of the training data. This approach significantly enhances feature projection to the target space compared to alternative methods. Moreover, it offers notable advantages in terms of memory and time efficiency, particularly crucial for handling extensive 3D volumes. 
    \item We conducted extensive experiments comparing our model with SOTA 2D and 3D methods, demonstrating that our model achieves SOTA performance in terms of accuracy, memory usuage and time consumption. Additionally, we evaluated the performance by individually removing each component from our model, quantitatively demonstrating the significance of each component.
\end{enumerate}

\section{Material and Methods}
\subsection{Dataset}
\begin{figure}[!ht]
\centering
\includegraphics[width=120mm]{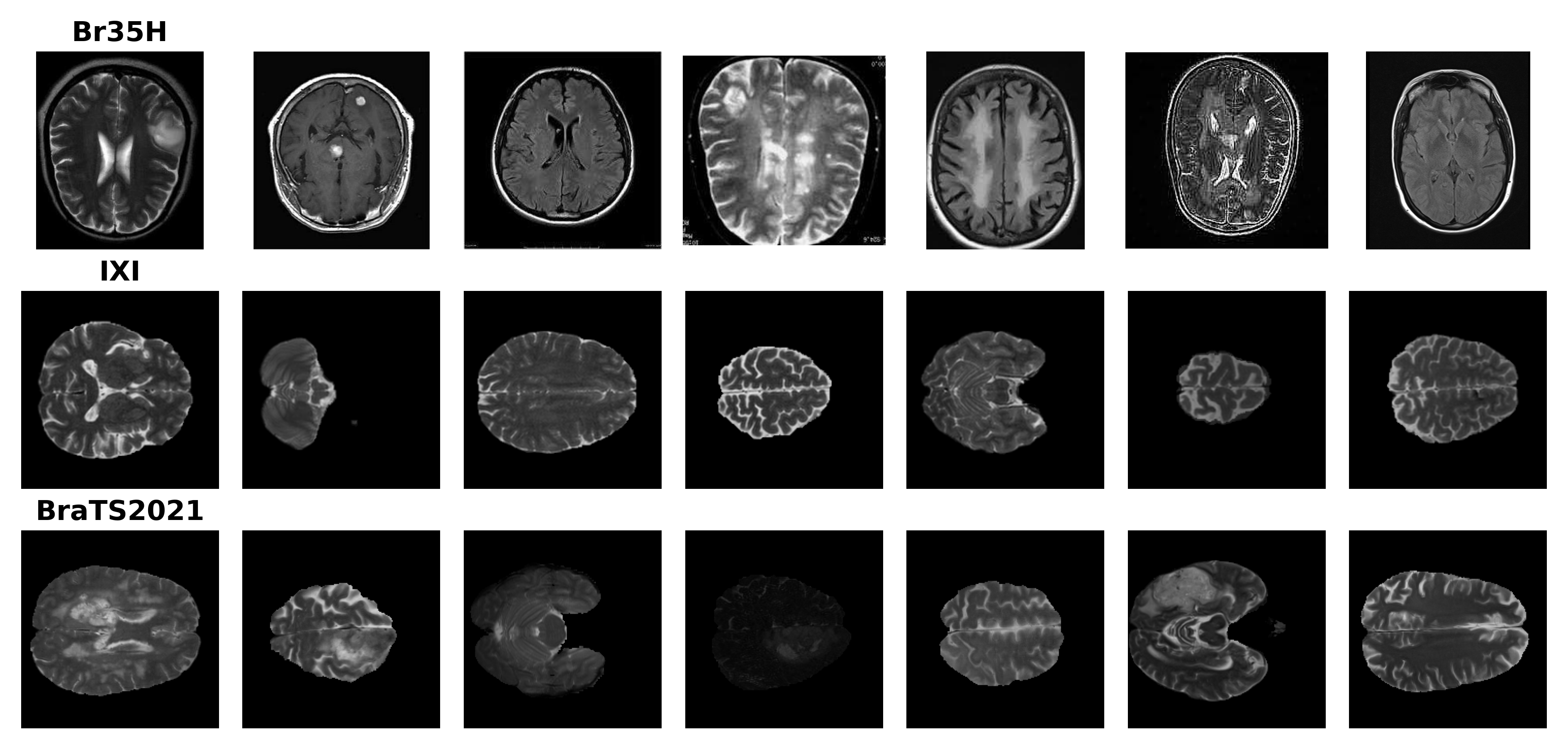}
\caption{{\bf Samples from datasets} This figure displays samples from each dataset used in this study. Each row represents samples from a different dataset, providing a visual comparison of their characteristics and preprocessing differences. Samples from the IXI and BraTS2021 shown in this figure have undergone preprocessing.
}

\label{fig:datasets}
\end{figure}

\textbf{Br35H} is a 2D brain MRI dataset on Kaggle \cite{br35h}. It contains 1,500 2-D image slices that are tumorous and 1,500 images that are non-tumorous. We resized the non-tumorous images to 256x256 and used them to fine tune the pre-trained model. From Fig~\ref{fig:datasets}, we observe that images from Br35H are inconsistent, not skull-stripped, and represent only the central area. The \textbf{IXI} dataset \cite{IXI} contains 576 healthy subjects from multiple institutions. The spatial resolution of all slices is $0.94 \times 0.94 \times 1.25$ $mm^3$, with an in-plane matrix size fixed at $256 \times 256$, and the number of slice ranged from 28 to 136. We use the T2-weighted volumes for training. The \textbf{BraTS2021} dataset  \cite{baid2021rsnabrats,bakas2017advancingbrats,menze2014multimodalbrats} comprises a total of 1,251 subjects with Glioblastoma. These images are interpolated to $1 \times 1 \times 1$ $mm^3$ resolution, skull-stripped, and finally cropped to a fixed size of $240 \times 240 \times 155$. We use the T2-weighted volumes for testing. Within this dataset, 251 volumes are designated for validation and 1,000 are used for testing. Samples from these three datasets are shown in Fig~\ref{fig:datasets}
% Place figure captions after the first paragraph in which they are cited.

\subsection{Pre-processing}
We followed the 3D brain MRI volume preprocessing procedures in \cite{luo2023uad3dbrain}, used HD-BET tool \cite{isensee2019hdbet} to skull-stripped the T2 volumes from IXI, and co-registered all the volumes from IXI and BraTS2021 to the Montreal Neurological Institute (MNI) template using the flirt tool of FMRIB software library \cite{jenkinson2001global,jenkinson2002improved}. 

\subsection{Anomaly Scores}
For our model, the log-likelihood of the distribution can be consider as the normality score, and thus the pixel-level anomaly score is defined as 
\begin{eqnarray}
    s(x) = 1-\exp{(\log{p(x)})},
\end{eqnarray} 
where $s(x)$ represents the anomaly score for pixel $x$. The image-level anomaly score is obtained as the maximum pixel anomaly score across the image. Given the extensive volumes in the testing dataset, we evaluate the metrics at the pixel level for each volume. This process involves normalizing the pixel-level anomaly score within each volume, calculating the metrics for each volume, and then computing the mean of these metrics across all volumes to obtain the final results. For image (slice)-level evaluation, we normalize and calculate the metrics across the entire testing dataset.

\subsection{Evaluation Metrics}
\label{sec:metrics}
In our paper, we utilize a uniform set of metrics to evaluate the performance of our model at both the image level and pixel level. We assess using Area Under the Receiver Operating Characteristic Curve (AUROC), Area Under the Precision-Recall Curve (AUPRC), accuracy (ACC), specificity (Spec), precision (Prec), F1 score (F1), and Maximum Dice Score $\lceil \text{DICE} \rceil$. These metrics allow us to comprehensively measure the model's ability to distinguish between positive and negative instances, balance precision and recall, and achieve high accuracy across different thresholds. To optimize performance evaluation, we dynamically determine the F1 threshold that maximizes F1 score, striking a balance between precision and recall, thus offering an optimal threshold. We use this F1 threshold to achieve the best accuracy, specificity, precision and F1 score.

The AUROC quantifies the model's capacity to differentiate between classes under varying thresholds, offering a broad view of classification performance. AUPRC is particularly valuable in scenarios with class imbalances, as it focuses on the precision-recall trade-off, providing insights into the model's effectiveness in identifying true positives amidst a large number of false positives.

Specificity and precision highlight the model’s accuracy in identifying true negatives and true positives, respectively, while the F1 score and Maximum Dice Score offer insights into the harmonic mean of precision and recall, and the spatial overlap accuracy between predicted and actual regions. The Maximum Dice Score allows for the assessment of the model's optimal performance under the best possible conditions. It is a critical measure in fields like medical imaging, where precise segmentation of anatomical structures is essential for accurate diagnosis and treatment planning.

For the pixel-level evaluation, all the aforementioned metrics apply, with the addition of the per-region-overlay (PRO) metric, which evaluates the spatial correspondence between predicted and ground truth regions, enhancing our understanding of the model’s localization accuracy. Unlike the image-level evaluation, PRO specifically helps in assessing the effectiveness of our model in segmenting and localizing regions at the pixel level.

We observe that the $\lceil \text{DICE} \rceil$ score is exactly the same as the maximum F1 score. This similarity arises because both metrics assess the balance between precision and recall in binary classification and segmentation tasks. Therefore, for the results presented in this paper, we use only the $\lceil \text{DICE} \rceil$ score.

\subsection{Network architecture and training setup}
\begin{figure}[!ht]
\centering
\includegraphics[width=130mm]{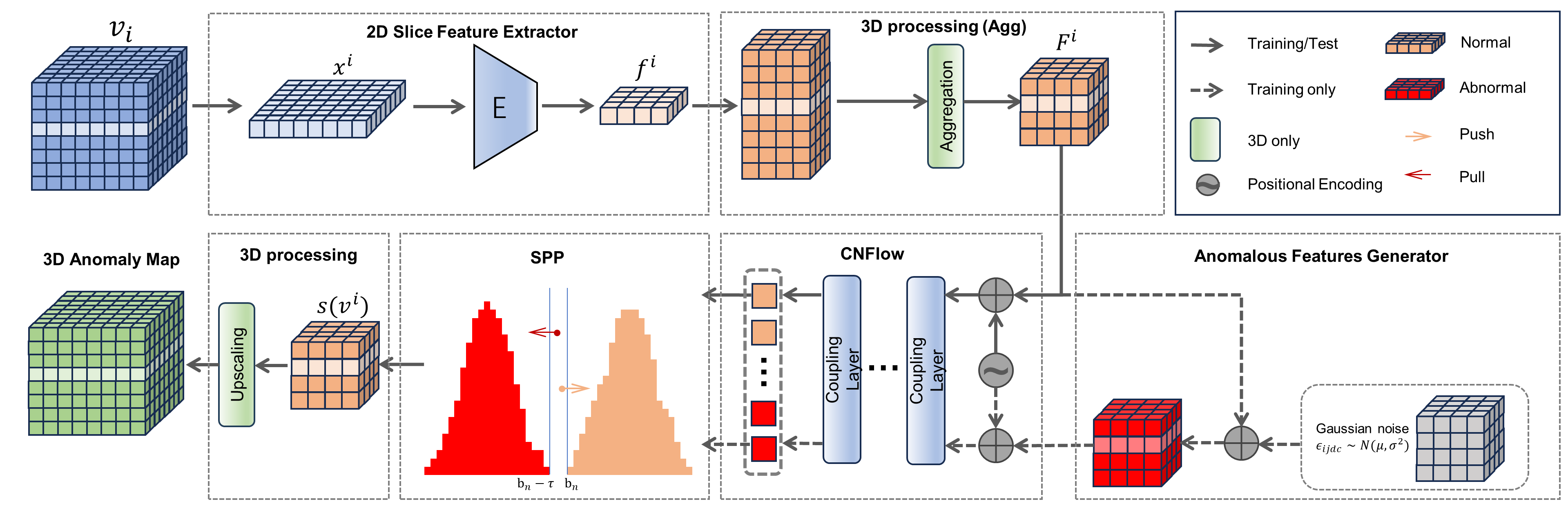}
\caption{{\bf SimpleSliceNet overall architecture}
Overview of the proposed SimpleSliceNet. The slice encoder $\boldsymbol{E}$ utilizes a backbone pre-trained on ImageNet to extract multi-layer feature maps $\boldsymbol{f_i}$ at low resolution. Anomalies are synthesized by introducing Gaussian noise into the feature space. Subsequently, these features are fed into \textbf{CNF} to estimate the log likelihood of the normal and anomalous distribution, and employ a \textbf{SPP} loss to refine the explicit boundaries of the normal distribution. Our SimpleSliceNet extracts slice features using the slice encoder and aggregates the resulting slice features via permutation invariant operations to achieve the features $\boldsymbol{F^i}$ of the volume $\boldsymbol{v^i}$ in low resolution. We process the feature vector of each voxel exactly like what we do on 2D slices to obtain anomaly map $\boldsymbol{s(v^i)}$. We finally upscale the anomaly map to the same resolution as that of the original volume.}
\label{fig:model}
\end{figure}

Our 2D slice encoder consists of a neural network backbone pre-trained on ImageNet, and conditional normalizing flow. The 3D brain MRI volume is processed by extracting slice features using 2D slice encoder and aggregating them to utilize depth-wise information. Fig~\ref{fig:model} shows the detail of our model architecture.

For all purpose, we use the WideResnet50 pre-trained on ImageNet \cite{deng2009imagenet} as the backbone which is further fine-tuned using EDC. More detail about the backbone model and its fine-tuning is presented in the \nameref{sec:edc}. Through an ablation study in \cite{liu2023simplenet} and our own experiments, we determined that utilizing features from 2nd and 3rd intermediate layers yields the best performance. The anomalies are synthesized by adding i.i.d Gaussian noise $\mathbf{N}(0,\sigma^2)$, where $\sigma$ is to 0.06 BraTS experiments. In the condition normalizing flow architecture, we employ 8 coupling layers, with each layer consisting of two linear layers and one ReLU activation layer. The target embedding dimension is set to 1024. The parameters of the CNF are optimized using an Adam optimizer with a learning rate to 0.001. For fair comparison, the batch size of all 3D models are 1 (volume), and the batch size of all 2D models are 96 slices (one volume).

% Place figure captions after the first paragraph in which they are cited.

\subsection{Slice Feature Extractor}\label{subsec:slice}
We employ a model pre-trained on ImageNet to serve as a feature extractor, extracting multi-layer features which is used in \cite{liu2023simplenet,roth2022towards}. Anomalies are generated by introducing Gaussian noise to these extracted features inspired by \cite{liu2023simplenet}. 

We denote training and testing sets as $I_{train}$ and $I_{test}$, respectively. The training dataset ($I_{train}$) contains only normal samples and the testing dataset ($I_{test}$) contains a mixture of normal and abnormal samples.
%Considering we have the training dataset and testing dataset as $I_{train}$ and $I_{test}$ respectively, where training dataset contains only normal samples and the testing dataset contains a mixture of normal and abnormal samples. 
Each image from the dataset is defined as $x_i \in I_{train} \cup I_{test}$ where $x_i \in \mathbb{R}^{3 \times H \times W}$. We passed in the samples into a neural network backbone $B$ to extract features. Then we define the output from the intermediate layer $l \in L$ of the backbone as $x^{l,i} \sim B^l(x_i)$, where $x^{l,i} \in \mathbb{R}^{C_l \times H_l \times W_l}$ and $(C_l,\ H_l,\ W_l)$ represents the dimensions of the features map from layer $l$. Let's define $x_{h,w}^{l,i}$ as the feature representation at location (h,w). Since we try to utilize the local environment information, we aggregate the neighborhood with a patch of size $p$ to obtain a local feature vector. The local feature vector of image $x^i$ from layer $l$ at location $(h,w)$ is defined as
\begin{equation}
      z_{h,w}^{l,i} = Aggregate(\{x_{h,w}^{l,i} | h'\in [h-\lfloor p/2 \rfloor, \dots, h+\lfloor p/2 \rfloor],
      w' \in [w-\lfloor p/2 \rfloor, \dots, w+\lfloor p/2 \rfloor]\}).
\end{equation}

Then, we upscale all the feature maps with different resolution to same higher resolution and concatenate them together. In our case, we choose the dimension size of the first layer $(H_0,W_0)$. Then the extracted map of each sample $f^i \in \mathbb{R}^{H_0,W_0,C}$ is defined as
\begin{equation}
        f^i = concat({upscaling(z^{l,i},(H_0,W_0))|l \in L}),
\end{equation}
where $C = \sum_{l \in L}C^l$. We further aggregate the channel information to decrease the channel size to a desired dimension. For convenience and further definition of 3D formulation, we simplify the feature extraction process as 
\begin{equation}
    f^i = E(x^i).
\end{equation}

\subsection{2D-Slice-Model and 3D Aggregation}
Define a brain MRI volume from a dataset $V$ as $v^i \in \mathbb{R}^{H \times W \times D}$, we extract the feature maps of each slice $v^i_j, j \in \{1,\dots,D\}$ from the volume $v^i$ using the 2D feature extractor $\mathbf{\textit{E}}$ mentioned in \nameref{subsec:slice}, so that $f^{i}_j = E(v^{i}_j)$. Then we aggregate the features of all slices to achieve the 3D anomaly representation of a volume. Then we define this process as
\begin{equation}  
    F^i = aggregate(\{E(v^i_j)|j \in [1,\dots,K] \}).
\end{equation}

\subsection{Conditional Normalizing Flow}
\label{sec:CNF}
Normalizing flow \cite{dinh2016densityNF} aims to represent the complex distributions with a simple one through a flow of successive invertible and differentiable transformations. The fundamental trick that makes normalizing flows work is the change of variables. The critical idea of the change of variables is transforming a random variable X into a tractable variable Z using an invertible and differential flow model $z=\phi(x)$. The log-likelihood of the random variable can be determined by
\begin{equation}
    \log{p_X(x)} = \log{p_Z(\phi(x))} + \log{\left|det(\frac{\,d \phi(x)}{\,d x})\right|}.
\end{equation}
To optimize the log-likelihood, it's equivalently to approximate the target distribution with a flow-based model with parameters $\Theta$ by minimizing the reverse KL distance
\begin{equation}
    \mathcal{L}(\Theta) = \mathbb{E}_{p_X(x;\Theta)}\left[\log{p_X(x;\Theta)}-\log{p_X(x)}\right] \label{eq:6}
    = -\mathbb{E}_{p_X(x;\Theta)}\left[\log{p_Z(\phi(x))} + \log{\left|det(\frac{\,d \phi(x)}{\,d x})\right|}\right].
\end{equation}
We followed \cite{yao2023BGAD,gudovskiy2022cflow}, assumed the simple tractable distribution obey the multivariate Gaussian distribution and rewritten the Eq~(\ref{eq:6}) as 
\begin{equation}
    \mathcal{L}(\Theta)  = -\mathbb{E}_{p_X(x;\Theta)}\left[\frac{d}{2}\log{2\pi} + \frac{1}{2}\phi(x)^T\phi(x) - \log{\left|det(\frac{\,d \phi(x)}{\,d x})\right|}\right] \label{eq:7}.
\end{equation}
For the transformation, we used coupling layers with fully connected layer which doesn't contain spatial information. In line with previous works \cite{yao2023BGAD,gudovskiy2022cflow}, we employ conditional normalizing flow to integrate 2D positional encoded information. 

\subsection{Encoder-Decoder Contrast}
\label{sec:edc}
%Comparing several ways to fit the pre-trained model to target medical dataset, including fine-tune and projection, we found that fine tuning the pre-trained model before training outperforms other two approaches and save training time. 
When utilizing pre-trained ImageNet models in the medical domain, one of the challenges is the disparity between the distribution of the ImageNet dataset and the target dataset. Common methods to address this issue typically involve fine-tuning the model or training additional layers to project the extracted features into the target space during training process. However, we are exploring an alternative approach to transfer knowledge across domains. We have discovered that fine-tuning the feature extractor before training outperforms other alternatives and can save training time in this case.
To keep the standard deviation and discriminative ability of the features from each intermediate layer. We fine-tune the pre-trained model using stop-gradient and global cosine distance proposed by \cite{guo2023EDC}. To maintain the assumption that pre-trained model has no knowledge on the training distribution, we fine-tune the model on an independent 2D MRI images dataset called Br25H \cite{br35h}. To fine-tune the pre-trained model on target dataset, \cite{guo2023EDC} create a decoder with a reverse architecture of the pre-trained model to reconstruct the images using encoder-decoder framework. For each l-th intermediate layer where $l \in L$, they denote $f_E^l, f_D^l \in \mathbb{R}^{H_l,W_l, C_l}$ as the features maps from the l-th layer of the encoder and decoder respectively, where $(H_l, W_l, C_l)$ is the dimension of the feature map. 
Then the reconstruction residuals are minimized by the global cosine distance loss
\begin{equation}
    \mathcal{L}_{global} = \sum_{l=1}^L 1-\frac{sg(F(f_E^l)^T) \cdot F(f_D^L)}{sg(\Vert F(f_E^l)\Vert)\Vert F(f_D^l)\Vert},
\end{equation}
where $sg$ represents the stop-gradient operation, and $F$ denote a flatten operation that casts the 2D feature map into feature vector. The stop-gradient operation won't affect the forward process and will consider the encoder parameters as constant while optimized them through the gradient from the decoder. Authors of \cite{guo2023EDC} proved that the stop-gradient operation can help to find discriminative features and the global cosine distance can address the instability problem.

\subsection{Semi-push-pull Paradigm}
Using discriminator to classify the normal features and noisy features (synthesized anomalies) may cause some false positive. We utilized a Semi-Push-Pull (SPP) loss to softly push the boundaries of normal features. We evaluated three candidate losses: Pair-wise Ranking Loss (PRL) \cite{chopra2005learning}, Triplet Loss (TL) \cite{schroff2015facenet}, and Boundary-guided Semi-Push-Pull Loss (BG-SPP) \cite{yao2023BGAD}. We grouped these losses under the Semi-Push-Pull Paradigm because they similarly pull normal features together and softly push anomalous features away with a margin. PRL and TL, two well-known losses, enforce a margin between positive and negative pairs, ensuring that similar instances are brought closer together in the feature space than dissimilar ones. Here, we apply these loss functions directly to the log probabilities returned by the CNF instead of to the feature embeddings. The BG-SPP loss is defined as 
\begin{equation}
    \mathcal{L}_{bg-spp} = \sum_{i=1}^N \left|min((\log{p_i}-b_n),0)\right| +\sum_{j=1}^M \left|max((\log{p_j}-b_n+\tau),0)\right| \label{eq:9},
\end{equation} 
where $b_n$ is the normal boundary obtained by selecting the $\beta$-th percentile of the sorted normal log-likelihood distribution, the $\tau$ is the margin between the boundaries of the normal and abnormal features, and $\log{p_i}$ is the log likelihood output from the CNF in \nameref{sec:CNF}. Then the final objective function would be to maximize the log-likelihood of the normal samples Eq~(\ref{eq:7}) and to minimize the SPP loss. The final loss function is
\begin{equation}
    \mathcal{L}oss = \mathcal{L}(\Theta) + \mathcal{L}_{spp}.
\end{equation}

\section{Results}
In this study, we used the healthy subjects from IXI for training and unhealthy brains from BraTS2021 for validation and testing. The pre-processed volumes from both datasets were center-cropped to $192\times 192 \times 96$ and the intensity range was rescaled to [0,1]. Our model was benchmarked against the latest 3D models specifically designed for brain MRI data. Additionally, we explored a straightforward approach of processing brain MRI volumes slice-by-slice using 2D methods, allowing us to also compare our model against SOTA 2D methods. Among these, \textbf{EDC} \cite{guo2023EDC} and \textbf{DAE} \cite{kascenas2023role} are 2D reconstruction-based methods, while \textbf{3D-AE} \cite{luo2023uad3dbrain} is a 3D reconstruction-based method tailored for brain MRI data. To achieve a more comparative evaluation and comprehensive study across a broader range of methods, we included comparisons with SOTA methods used in industrial datasets to access their potential in the medical field. Representative models were chosen for benchmarking against our method, including \textbf{SimpleNet} \cite{liu2023simplenet}, \textbf{BGAD} \cite{yao2023BGAD} and \textbf{CFLOW\_AD} \cite{gudovskiy2022cflow} for embedding-based methods; \textbf{AE\_FLOW} \cite{zhao2022aeflow} for reconstruction-based methods; \textbf{CutPaste} \cite{li2021cutpaste} for synthesis-based methods; \textbf{PatchCore} \cite{roth2022towards} for memory-bank methods; and \textbf{RD++} \cite{tien2023revisitingRST} for knowledge-distillation based methods. \cite{yao2023BGAD} is a semi-supervised method with a mechanism similar to our model, as both utilize conditional normalizing flow and a SPP loss. Therefore, we selected it as a candidate for comparison. $\text{BGAD}^\text{w/o}$ refers to the unsupervised version of BGAD, which operates without seen anomalies and BG-SPP loss. $\text{BGAD}^\text{100}$ denotes the semi-supervised version with 100 known anomalous slices.

\subsection{Localization Quantitative Results}
Table~\ref{tab:p_metrics} displays the results of pixel-level metrics for all the models. According to the results, our method achieves the best performance, while \textbf{PatchCore} and \textbf{RD++} show competitive performance in terms of AUROC. However, there is a significant improvement on AUPRC, PRO and $\lceil \text{DICE} \rceil$ compared to these methods. As discussed in \nameref{sec:metrics}, these three metrics are particularly crucial and representative in tasks involving imbalanced anomaly segmentation.
% Place tables after the first paragraph in which they are cited.
\begin{table}[!ht]
%\begin{adjustwidth}{-2.25in}{0in} % Comment out/remove adjustwidth environment if table fits in text column.
\centering
\caption{
{\bf Quantitative results at pixel level for all models} This table presents the Quantitative results of all models at the pixel level, providing a comprehensive overview of each model's effectiveness in anomaly localization. \textit{Ours} denotes our method variant with Triplet Loss.}
\begin{tabular}{|l|l|l|l|l|l|l|l|}
\hline
%\multicolumn{7}{|l|}{\bf Heading1} & \multicolumn{4}{|l|}{\bf Heading2}\\ \thickhline
& AUROC & AUPRC & PRO & $\lceil \text{DICE} \rceil$ & Spec & ACC & Prec \\ \hline
SimpleNet & 86.9 & 23.31 & 61.05 & 28.06 & 24.27 & 91.84 & 93.09 \\ 
$\text{BGAD}^{\text{w/o}}$ & 77.88 & 7.82 & 37.81 & 13.4 & 8.82 & 79.09 & 79.7 \\
$\text{BGAD}^{\text{100}}$ &92.04 & 35.37 & 73.13 & 37.79 & 35.26 & 95&96.24\\ 
CFLOW\_AD &83.21 & 13.29 & 46.34 & 19.87 & 14.3 & 82.68 & 83.44  \\ \hline
EDC & 88.77 & 32.86 & 68.84 & 37.74 & 34.6 & 90.57 & 91.35  \\
DAE & 87.89 & 16.30 & 75.08 & 24.88 & 19.95 & 89.40 & 90.42  \\ 
AE\_FLOW & 82.28 & 12.91 & 68.85 & 19.84 & 14.53 & 81.53 & 82.12 \\ \hline
Cut\_Paste & 73.15 & 3.64 & 28.53 & 7.29 & 4.5 & 87.21 & 88.12  \\ \hline
PatchCore & 95.51 & 44.75 & 76.44 & 48.52 & 44.26 & 96.52 & \textbf{97.42}  \\ \hline
RD++ & 95.25 & 38.50 & 74.99 & 44.07 & 37.99 & \textbf{95.75} & 96.64  \\ \hline
3D\_AE & 86.69 & 18.03 & 73.42 & 25.79 & 20.01 & 88.37 & 89.19  \\ \hline
Ours & \textbf{95.55} & \textbf{51.33} & \textbf{76.98} & \textbf{51.07} & \textbf{48.5} & 95.67 & 96.48 \\ \hline
\end{tabular}
%\begin{flushleft}  
%\end{flushleft}
\label{tab:p_metrics}
%\end{adjustwidth}
\end{table}

\subsection{Detection Quantitative Results}
Table~\ref{tab:I_metrics} presents the quantitative results at the image level for all models. Our methods generally outperforms the other SOTA methods, except for \textbf{EDC}, which achieves the best performance across all metrics. However, despite its superior image-level metrics, \textbf{EDC} demonstrates poor performance in localization ad lacks stability, as indicated by the numerical results on localization and the qualitative assessments discussed in other subsections. Furthermore, although our model shows visually superior tumor detection compared to \textbf{EDC}, the lower image-level may be attributed to our spatial anomaly detection at lower resolutions, which occasionally results in misclassification of slices due to imprecise tumor boundary detection. Nevertheless, when evaluating the volume as a whole rather than individual slices, our model effectively detects spatial anomalies, showcasing its robustness in comprehensive anomaly detection.

% Place tables after the first paragraph in which they are cited.
\begin{table}[!ht]
%\begin{adjustwidth}{-2.25in}{0in} % Comment out/remove adjustwidth environment if table fits in text column.
\centering
\caption{
{\bf Quantitative results at the image level of all models} This table presents the Quantitative results of all models at the image level, providing a comprehensive overview of each model's effectiveness in anomaly detection. \textit{Ours} denotes our method variant with Triplet Loss.}
\begin{tabular}{|l|l|l|l|l|l|l|}
\hline
%\multicolumn{7}{|l|}{\bf Heading1} & \multicolumn{4}{|l|}{\bf Heading2}\\ \thickhline
& AUROC & AUPRC & $\lceil \text{DICE} \rceil$ & Spec & ACC & Prec\\ \hline
SimpleNet & 62.3 & 73.12 & 78.87 & 65.82 & 65.81 & 5.69\\ 
$\text{BGAD}^{\text{w/o}}$ & 51.93 & 65.53 & 78.68 & 64.86 & 64.86 & 0 \\
$\text{BGAD}^{\text{100}}$ & 66.44 & 78.21 & 78.7 & 64.88 & 64.89 & 0.15\\ 
CFLOW\_AD & 51.02 & 64.35 & 78.03 & 63.98 &63.98& 0.09\\ \hline
EDC & \textbf{79.82} & \textbf{86.56} & \textbf{81.45} & \textbf{76.46} & \textbf{74.26} & \textbf{50.48}\\
DAE & 67.85 & 77.87 & 78.86 & 66.92 & 66.62 & 12.42 \\ 
AE\_FLOW & 63.74 & 73.12 & 78.72 & 65.3 & 65.26 & 2.84 \\ \hline
Cut\_Paste & 55.77 & 71.04 & 78.73 & 64.92 & 64.92 & 0  \\ \hline
PatchCore & 71.1 & 78.82 & 79.1 & 68.03 & 67.79 & 19.26  \\ \hline
RD++ & 74.57 & 81.2 & 79.94 & 70.34 & 70.02 & 29.01  \\ \hline
3D\_AE & 69.91 & 79.47 & 79.44 & 68.43 & 68.22 & 19.42 \\ \hline
Ours & 75.31 & 84.57 & 79.5 & 69.05 & 68.67 & 22.48 \\ \hline
\end{tabular}
%\begin{flushleft} 
%\end{flushleft}
\label{tab:I_metrics}
%\end{adjustwidth}
\end{table}

\subsection{Localization Qualitative Results}
\begin{figure}[!ht]
\centering
\includegraphics[width=130mm]{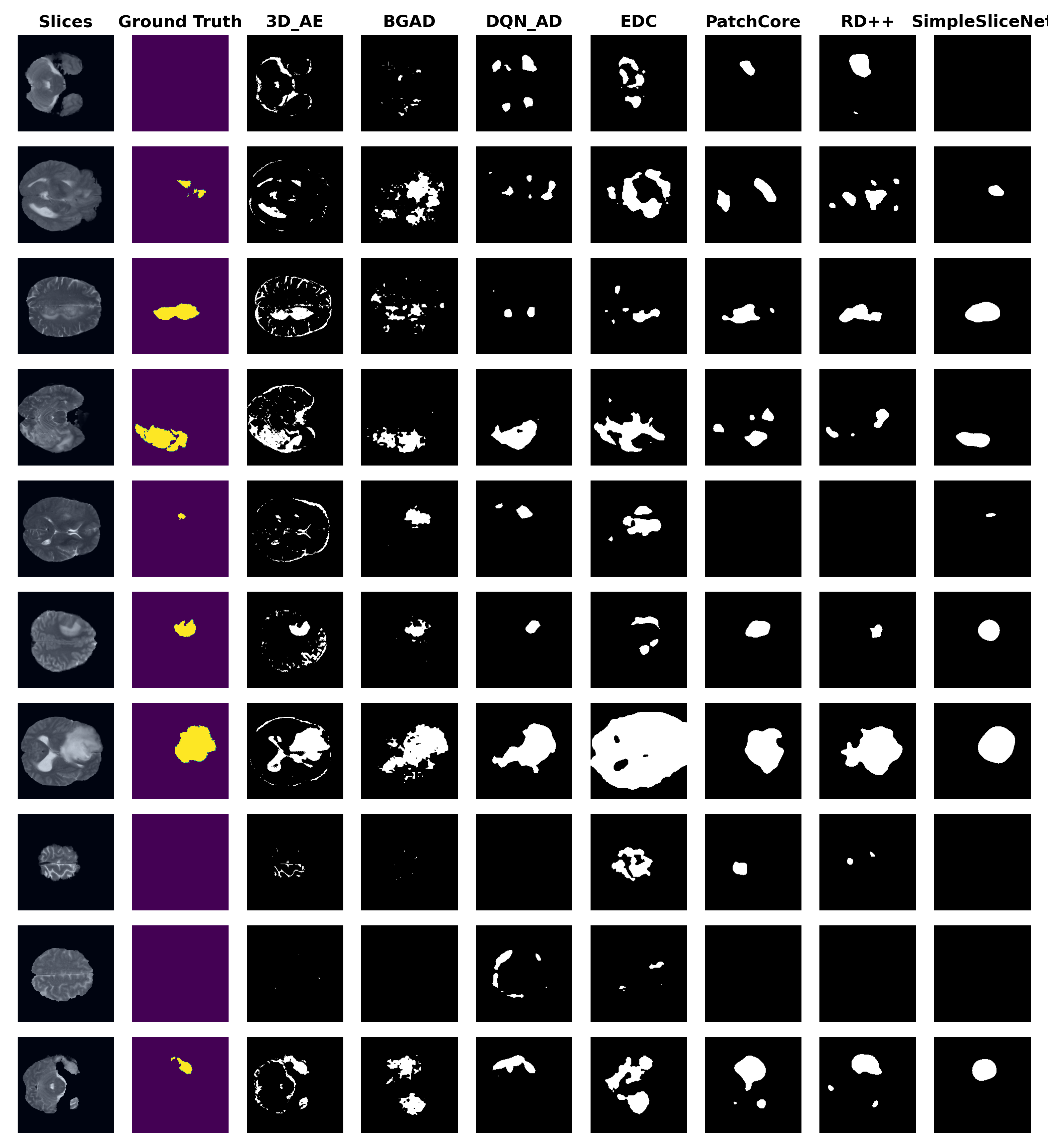}
\caption{{\bf Qualitative Results on the SOTA methods} This figure displays slices from volumes selected from the testing dataset, chosen to represent different levels within the volume and varied visual structures. This selection ensures a more comprehensive evaluation of the models' performance. The first column presents the images (slices) from brain MRI volumes and the second column illustrates the ground-truth data for anomalous regions. The subsequent columns depict the binary anomaly map predicted by selected models. These maps are generated by applying the F1 threshold, which yield the best F1 score, to convert anomaly scores into binary masks.
}
\label{fig:qualitative}
\end{figure}

Fig~\ref{fig:qualitative} presents the qualitative result for each model. The 2D and 3D reconstruction-based methods, \textbf{EDC} and \textbf{3D\_AE} generate a significant number of false positives. Specifically, \textbf{EDC} often misses the central area of the anomaly, likely because this region is easier to reconstruct, whereas the transitional areas between normal and abnormal tissues pose more challenges to reconstruct. \textbf{BGAD}, while performing better, also generates some small false positives and the detected anomalies are not well-concentrated. \textbf{PatchCore} and \textbf{RD++} show improved performance but still produce some false positive regions. In contrast, our model accurately localizes that the anomalies without generating false positives away from the anomalous regions. Furthermore, the detected anomalous areas in our model encompass more true abnormal regions than those identified by other methods.

\begin{figure}[!ht]
\centering
\includegraphics[width=130mm]{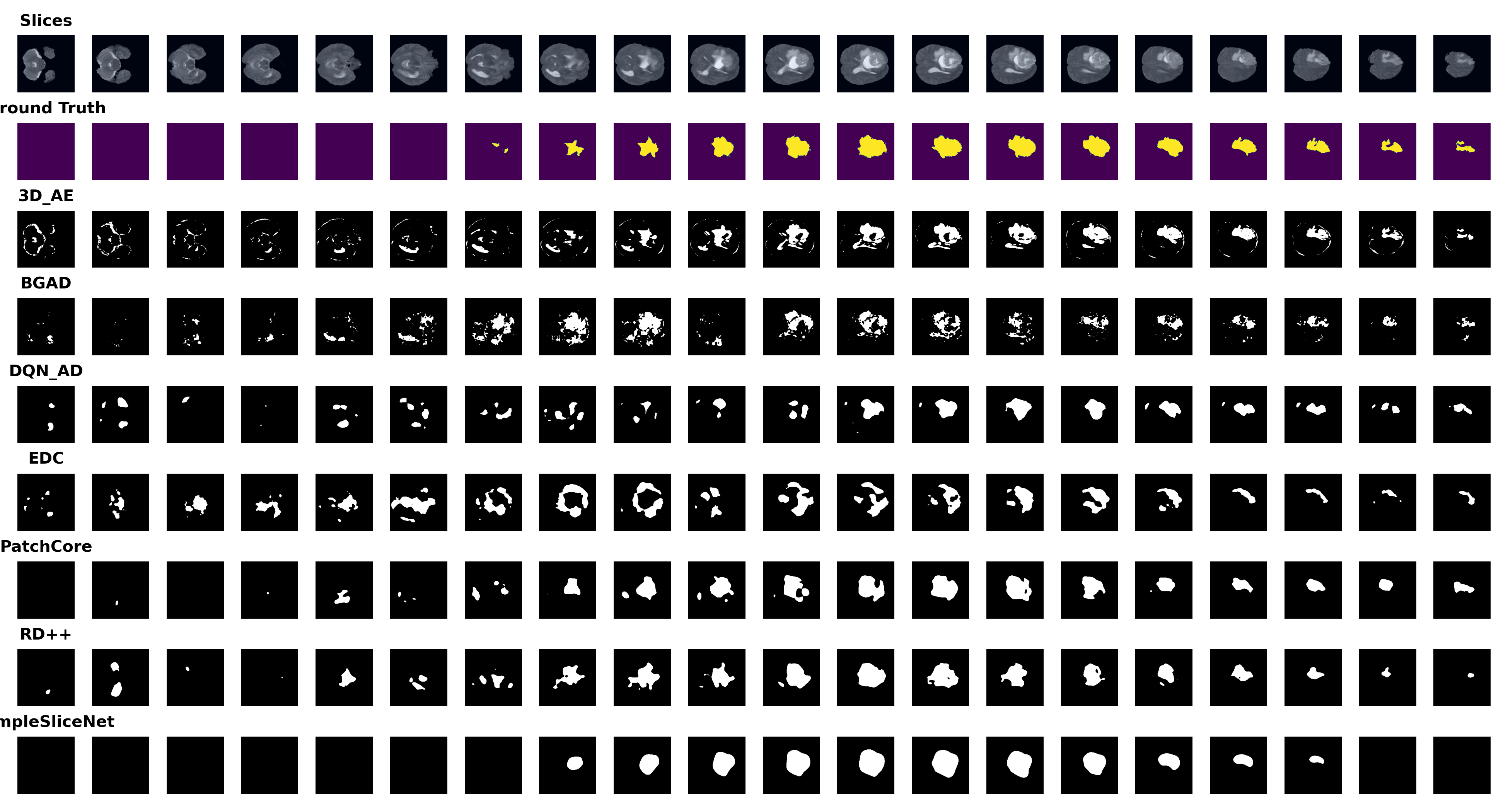}
\caption{{\bf Qualitative Results on the SOTA methods} This figure displays intermediate slices in sequence from a single volume selected from the testing dataset. The visualization is designed to demonstrate how well the model performs spatially across an entire volume. The first row presents the images (slices) from brain MRI volumes and the second row shows the ground-truth for the anomalous regions. The remaining rows depict the binary anomaly maps predicted by SOTA models. These maps are generated by applying the F1 threshold, which achieves the best F1 score, to convert anomaly scores into binary masks.
}
\label{fig:qualitative_volume}
\end{figure}

Fig~\ref{fig:qualitative_volume} displays the qualitative results with slices taken sequentially from a single volume. This visualization aids in assessing the spatial localization of anomalies. The other methods tend to generate small false positive regions, particularly in normal slices where anomalies slices where anomalies are erroneously detected. In contrast our method effectively localizes spatial anomalies without generating unnecessary false positives, leading to a more concise and straightforward predicted anomaly map. However, this focus on precise anomaly positioning and minimizing false positives compromises the ability to accurately delineate the anomaly's boundaries. Consequently, the detected regions appear rough, occasionally missing some areas along the depth axis and misclassifying some slices. This limitation likely contributes to our method's lower image-level metrics compared to those of \textbf{EDC}.

% Place figure captions after the first paragraph in which they are cited.

\subsection{Training Process}
\begin{figure}[!ht]
\centering
\includegraphics[width=140mm]{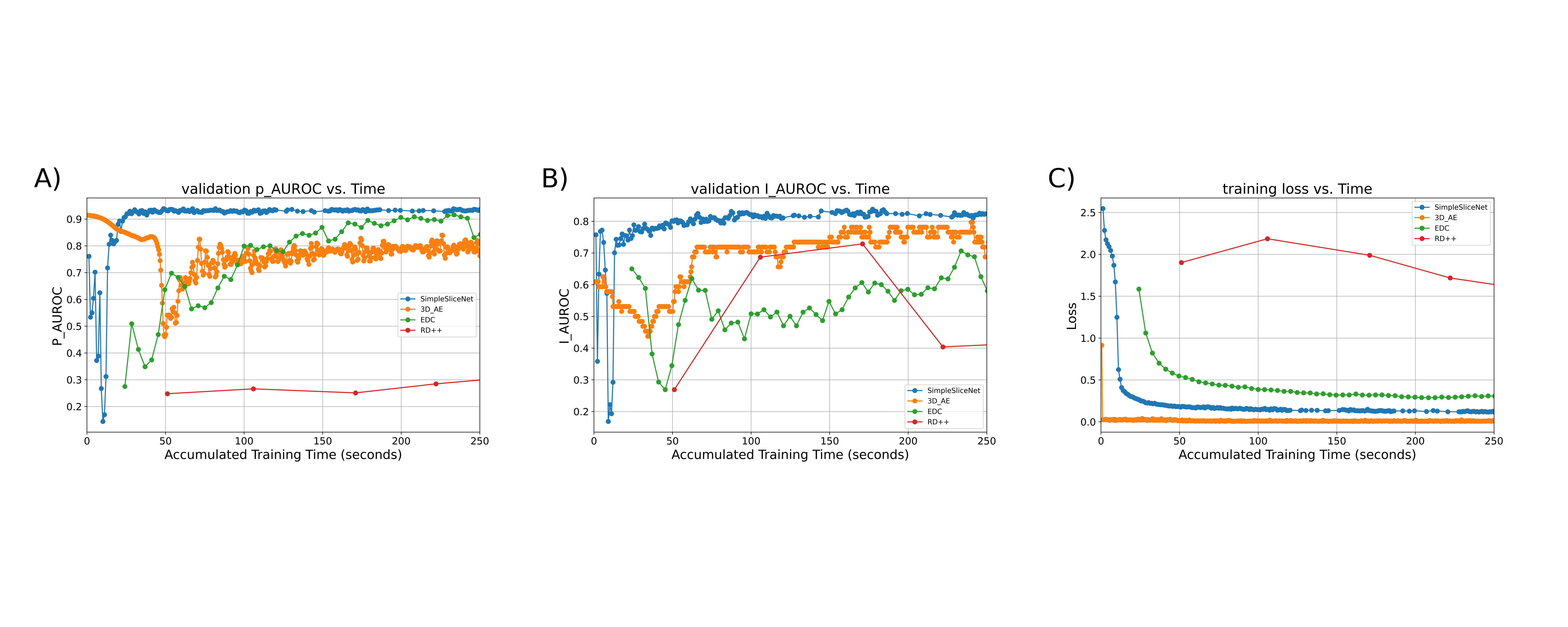}
\caption{{\bf Plots of AUROC and loss during training} This figure presents plots of image-level and pixel-level AUROC, along with average training loss, observed during the training phase. For practicality, we trained on 60 volumes from the training dataset and used 20 volumes from the validation dataset for evaluation. We assessed model performance after processing each volume (every single iteration), and the results for the first 250 seconds are plotted. The interval between two consecutive points on a line indicates the time required for the model to process one volume, consisting of 96 slices. A: plot of pixel-level AUROC. B: plot of image-level AUROC. C: plot of average training loss. 
}
\label{fig:plots}
\end{figure}

Fig~\ref{fig:plots} illustrates that RD++ requires the longest time to process a brain MRI volume and to converge. It's evident from the plot that our method converges more quickly and demonstrates greater stability compared to the other three methods evaluated.
% Place figure captions after the first paragraph in which they are cited.

\subsection{Computational Cost}
In Table~\ref{tab:memory}, we compare those well-performing models in terms of memory size and inference time. This table shows that our methods has the smallest number of trainable parameters, which contributes to a shorter convergence time. Additionally, it requires less estimated memory and has a shorter inference time. While our inference time is slower than that of \textbf{EDC} and our estimated memory size is smaller than that of \textbf{PatchCore}, the differences are marginal. Moreover, \textbf{PatchCore} necessitates extensive additional memory to store the memory bank and a longer inference time to search for the nearest neighbors within the bank. \textbf{EDC}, while faster, shows inferior performance in localization tasks and has a larger number of trainable parameters.
% Place tables after the first paragraph in which they are cited.
\begin{table}[!ht]
%\begin{adjustwidth}{-2.25in}{0in} % Comment out/remove adjustwidth environment if table fits in text column.
\centering
\caption{
{\bf Estimate memory size and inference time} \textit{Trainable} refers to the number of parameters in the model that can be updated during training, expressed in millions. \textit{Non-trainable} denotes the number of parameters that remain fixed during training, also in millions. \textit{Memory Size} indicates the estimated amount of memory required to store both trainable and non-trainable parameters, measured in megabytes. \textit{inference time} describes the time needed for the model to compute the anomaly map for a volume comprising 96 slices, with the time measured in seconds.}
\begin{tabular}{|l|l|l|l|l|}
\hline
& Trainable & Non-trainable & Memory Size & Inference time\\ \hline
%BGAD & 2.64m & 40.15m & 163.248mb & 0.35s \\ \hline
EDC & 137m & 0 & 525.58mb & 0.34s \\ \hline
PatchCore & 0 & 68.88m & 262.77mb & 14.57s\\ \hline
RD++ & 107.69m & 68.88m & 673.5mb & 11.79s \\ \hline
3D\_AE & 235.82m & 0 & 899.57mb & 1.25s \\ \hline
Ours & 3.2m & 69.93m & 269.41mb & 0.74s \\ \hline
\end{tabular}
%\begin{flushleft} 
%\end{flushleft}
\label{tab:memory}
%\end{adjustwidth}
\end{table}

\subsection{Ablation Study}
\subsubsection{Study on Semi-Push-Pull Losses}
Table~\ref{tab:SPP} illustrates the detection and localization performance associated with different SPP loss configurations. The comparison between the first three rows and the last row in this table indicates that incorporating an additional SPP loss significantly enhances performance. Among these three candidate losses, all show comparable performance, with the TL variant exhibiting a slight improvement.
% Place tables after the first paragraph in which they are cited.
\begin{table}[!ht]
%\begin{adjustwidth}{-2.25in}{0in} % Comment out/remove adjustwidth environment if table fits in text column.
\centering
\caption{
{\bf Quantitative results on different Semi-Push-Pull Loss} \textit{w/o SPP} refers to our model variant without the integration of any additional SPP loss. This configuration allows us to evaluate the impact of excluding the SPP loss on the models performance.}
\begin{tabular}{|l|l|l|l|l|l|l|l|}
\hline
& \multicolumn{3}{l|}{Image-level} & \multicolumn{4}{l|}{Pixel-level}\\ \cline{2-8}
& AUROC & AUPRC & $\lceil \text{DICE} \rceil$ & AUROC & AUPRC & PRO & $\lceil \text{DICE} \rceil$ \\ \hline
w/ TL & 75.31 & 84.57 & 79.5 & 95.55 & 51.33 & 76.98 & 51.07\\ \hline
w/ PRL & 74.08 & 83.95 & 79.29 & 94.94 & 47.35 & 75.13& 47.87\\ \hline
w/ BG-SPP & 73.8 & 82.67 & 79.67 & 95.8 & 53.32 & 75.83 & 52.8\\ \hline
w/o SPP & 69.27 & 80.34 & 79.15 & 92.43 & 34.8 & 67.6 & 38.46\\ \hline
\end{tabular}
%\begin{flushleft} 
%\end{flushleft}
\label{tab:SPP}
%\end{adjustwidth}
\end{table}

\subsubsection{Study on Components}
In Table~\ref{tab:components}, we investigate the significance of each component to address questions such as: Does aggregation on axial slices diminish information?  Does processing axial slices independently in the feature extractor limit the expressiveness of the resulting 3D feature volume? Can a pre-trained model tailored to brain MRI datasets enhance performance? Why is normalizing flow more effective than hard classification by a simple binary classification model? The table demonstrates that each modification significantly improves performance. Specifically, the comparison between the first and last row indicates that aggregation effectively utilizes axial information to enhance the model. However, it remains unclear whether the aggregated features are representative or contain irrelevant information, a common issue in embedding-based methods where feature extraction and anomaly detection are distinct processes. The comparison between the second and last row suggests that using a pre-trained model more attuned to brain MRI dataset is beneficial. For a fair comparison, we fine-tuned a pre-trained ImageNet model on an additional dataset of low quality and quantity to limit extraneous knowledge. We explored various methods to embed the extracted features into a space suitable for brain MRI dataset, including fine-tuning during training and adding a projection layer as done by SimpeNet \cite{liu2023simplenet}. Our modified pre-finetuned feature extractor with \textbf{EDC} acieved the best performance, stable training, and reduced training time. Additionally, we compare our model to a special case of SimpleNet with aggregation. This comparison further highlights the effectiveness of the CNF with SPP mechanism.

\begin{table}[!ht]
\tabcolsep=0.12cm
\centering
\caption{
{\bf Quantitative results of our method with different modifications} This table illustrates the performance of our model with the removal of each component, assessed using both pixel- and image-level metrics. \textit{Ours w/o Agg.} refers to our model without the part the slices' feature aggregating component, essentially transforming it into a 2D version that processes brain MRI data slice by slice. \textit{Ours w/o EDC} indicates that the feature extractor was not fine-tuned, with the extracted features directly applied to anomaly detection tasks. \textit{SimpleNet w/ Agg.} represents a modification where our CNF is replaced with a binary classification model, effectively making it a 3D version of SimpleNet that utilizes our aggregation approach. Finally, \textit{Ours} denotes our proposed method with Triplet Loss.}
\begin{tabular}{|l|l|l|l|l|l|l|l|}
\hline
& \multicolumn{3}{l|}{Image-level} & \multicolumn{4}{l|}{Pixel-level}\\ \cline{2-8}
& AUROC & AUPRC & $\lceil \text{DICE} \rceil$ & AUROC & AUPRC & PRO & $\lceil \text{DICE} \rceil$ \\ \hline
Ours w/o Agg. & 70.81 & 80.27 & 79.04 & 94.58 & 43.99 & 68.51&46.14 \\ \hline
Ours w/o EDC & 65.03 & 75.19 & 78.93 & 93.67 & 37.88 & 72.54 & 41.1 \\ \hline
SimpleNet w/ Agg. & 60.13 & 72.78 & 78.82 & 86.24 & 21.63 & 57.82 & 25.96 \\ \hline
Ours & 75.31 & 84.57 & 79.5 & 95.55 & 51.33 & 76.98 & 51.07 \\ \hline
\end{tabular}
%\begin{flushleft} 
%\end{flushleft}
\label{tab:components}
\end{table}

\section{Discussion}
In this study, we introduced SimpleSliceNet, an efficient framework designed to process both 2D and 3D brain MRI data for anomaly detection and localization. SimpleSliceNet leverages a backbone network to extract normal features and introduces Gaussian noise to these features to simulate anomalies. It utilizes a conditional normalizing flow combined with Triplet Loss to effectively delineate the boundaries of the normal distribution, adapting to the unique characteristics of brain MRI data. The quantitative detection and localization analyses presented in this paper offer valuable insights into the efficacy of different anomaly detection models in medical imaging, particularly in brain MRI. These findings highlight the strengths and limitations of our proposed model compared to other SOTA methods.

Unsupervised methods for images fall into three categories: reconstruction-based, synthesizing-based, and embedding-based. Reconstruction-based models are commonly used in brain MRI anomaly detection, while synthesizing-based and embedding-based approaches are less explored in this field.

% talk about reconstruction-based and their drawbacks
\textbf{Reconstruction-based} methods, including auto-encoder (AE) \cite{kascenas2022dae_mri,kascenas2023role} and Generative Adversarial Network (GAN) \cite{schlegl2019fanogan}, are extensively studied and considered SOTA in medical imaging. These models are trained solely on normal data, assuming anomalies cannot be accurately reconstructed, making the residual difference between input and reconstructed samples indicative of anomalies. For instance, in \cite{guo2023EDC}, the authors fine tuned a pre-trained network as an encoder and trained a decoder with a reversed architecture to reconstruct medical images, using contrastive loss to maintain the discriminative ability. Similarly, \cite{van2021ad_mri_chronic} used an AE framework to identify chronic brain infarcts on MRI, employing an additional decoder and discriminator for anomaly detection at the image level instead of directly taking the maximum anomaly scores on pixels. %U-Net \cite{ronneberger2015unet}, which incorporates skip-connection to alleviate the vanishing gradient problem and facilitate the training progress, has also been adopted for anomaly detection. As seen in \cite{zhang2023prototypical}, where a U-Net-like architecture detects anomalous patterns of varying sizes and scales. 
Some approaches, such as \cite{zhao2022aeflow,pinaya2022fastdiffusion}, utilize normalizing flow and diffusion models to compute anomaly likelihood on latent representations instead of pixel-level scores. Despite their utility, reconstruction-based models often produces blurry and low-quality images, and there is a trade-off between reconstruction ability and model generality, necessitating careful experimentation to balance these factors. Additionally, these methods generate noisy anomaly maps that require further post-processing steps. For medical data, the variation among normal samples (e.g. due to differences between patients or imaging infrastructures) may be greater than the variation of anomalous samples. In such cases, reconstruction methods that focus on deviations may not be effective and may create too many false positives. Our method aims to reduce false positives while maintaining performance. Qualitatively, while \textbf{EDC} and \textbf{3D\_AE} tend to generate a high number of positives and struggle with anomaly localization, our model consistently demonstrates superior accuracy in detecting true anomalies with minimal false positives. This precision is critical in clinical settings, correctly identifying anomalous regions can significantly impact diagnosis and treatment plans. However, the focus on minimizing false positives introduces challenges in precisely delineating the anomaly's boundaries, occasionally resulting in rough detected regions and misclassification of some slices. 

% talk about synthesizing-based and to introduce pros of our method
\textbf{Synthesizing-based} methods offer an alternative approach to anomaly detection by generating realistic anomalies from normal images. In \cite{li2021cutpaste}, small patches are cut from other normal images and pasted onto normal samples. To ensure smooth boundaries, NSA \cite{schluter2022NSA} integrated Poisson image editing to create smooth and continuous boundaries on synthesized anomalies. However, synthesized anomalies may not accurately represent real anomalies due to the unpredictable nature of anomalies within dataset, particular in tasks like brain MRI anomaly detection. 
%This is particularly challenging in tasks such as brain MRI anomaly detection, where synthesized anomalies struggle to closely resemble actual anomalies. 
We followed SimpleNet \cite{liu2023simplenet} to generate deviations within features to create anomalies, which may be more suitable for capturing a wider range of anomalies. %Moreover, anomalies in brain MRI data vary widely and can result from various abnormal conditions or unique circumstances during MRI acquisition. Learning from a small number of anomaly samples, particularly when targeting specific disease, may improve performance. Our SimpleSliceNet facilitates the integration of synthesizing-based methods or a small number of anomaly samples to enhance model performance.

\label{sec:embed}
% talk about embedding-based
\textbf{Embedding-based} methods involve learning the normal features of data and projecting them into an embedding space where abnormal features are further from normal ones. Many STOA models currently use pre-trained ImageNet models to extract meaningful features and save training time. SimpleNet \cite{liu2023simplenet} utilizes an ImageNet-pretrained encoder to extract multi-scale features, synthesize anomalies by adding Gaussian noise to the normal features, and trains a simple discriminator to classify the features. Normalizing Flow (NF) \cite{dinh2016densityNF} is used to convert the extracted features into a desired distribution, typically a Gaussian distribution. CFLOW-AD \cite{gudovskiy2022cflow} employs an ImageNet-pretrained encoder to extract features from intermediate layers and uses conditional NF (CNF) for each layer to obtain the normal distribution, enabling the detection of anomalous features at different scales. BGAD \cite{yao2023BGAD} aims to establish a clearer boundary on the normal distribution with a small number of anomalous examples through a soft pushing mechanism. These methods have achieved SOTA performance compared to other types of approach. Inspired by SimpleNet and BGAD, our method is specifically designed to address the challenges of medical data by generating synthetic anomalies at the feature level. This approach helps refine the normal boundary without disrupting the diverse normal distribution present in medical data.

% talk about knowledge distillaiton and their drawbacks and thus related to our motivation to use EDC
\textbf{Knowledge Distillation} methods  \cite{salehi2021multiresolution,zhang2023destseg,deng2022reverseST,tien2023revisitingRST} also utilize pre-trained models, training a Student network to learn the knowledge specific to normal data from a Teacher network, ensuring that the Student network only possesses knowledge about normal features. The residual between the output from Student and Teacher reveals anomaly information. However, these methods often demand significant memory and entail long training times. Additionally, for brain MRI data, there is a lack of suitable pre-trained models. To mitigate training workload and time, we still prefer models pre-trained on ImageNet to extract features, and investigate a better way to leverage pre-trained models on medical data.

% talk about 3D methods
For 3D brain MRI volumes, reconstruction-based models represent the SOTA unsupervised approach for detecting anomalies. \cite{luo2023uad3dbrain,pinaya2022uad3dbrain_transformer} used the idea of reconstruction to identify anomalies in these volumes. Instead of relying on the use of maximum pixel scores for detection in \cite{luo2023uad3dbrain}, \cite{pinaya2022uad3dbrain_transformer} used autoregressive transformers to analyze the distribution of the latent vector. Meanwhile, \cite{rudie2021UNETbrain,li2021UNETwhitematter} are supervised methods adopted 3D U-Net \cite{cciccek20163dunet} and 3D ResU-Net \cite{lee2017superhumanresunet} frameworks respectively to integrate feature maps from different resolution levels. However, these 3D CNN networks demand significant memory for parameter storage and considerable time for backpropagation. Instead, our model showcases computational efficiency. It requires fewer trainable parameters, operates with less memory, and achieves faster convergence. This efficiency does not compromise performance, which is pivotal for deploying these models in real-world clinical settings where processing speed and resource utilization are crucial.

Our model excels in several critical metrics such as AUROC, PRO, and $\lceil \text{DICE} \rceil$ metrics, which are essential for understanding the effectiveness of anomaly detection in imbalanced datasets. Despite the competitive performance of models like \textbf{PatchCore} and \textbf{RD++} on AUROC, our method's advancements in these specific metrics underscore its robustness and suitability for clinical applications that require precise anomaly delineation. Conversely, despite \textbf{EDC} showing outstanding performance across standard image-level metrics, its shortcomings in localization and stability - evidenced by both quantitative and qualitative evaluations - point to limitations in its practical applicability for consistent anomaly detection. Our model's superior tumor detection capability, though associated with lower image-level scores than \textbf{EDC}, can be attributed to our approach to spatial anomaly detection. While this approach reduces resolution to manage computational demands, it may lead to occasional misclassifications, especially at the boundaries of anomalous regions. However, they are significantly mitigated when considering the volumetric assessment of anomalies, where our model excel in detecting spatial discrepancies across the entire volume. 
These findings emphasize the need for future research to focus on enhancing resolution and boundary clarity in anomaly detection models without compromising on the computational efficiency required for real-time applications.

The component analysis reveals which aspects of the model architecture contribute most to its success and identifies areas needing refinement. A remaining question is whether the aggregated features representative for anomaly detection and if there are any other ways to better utilize depth-wise information. We conducted a preliminary experiment with depth-wise attention module, which showed promise result but required additional training time and computation resources. This area warrants further investigation. Our results also demonstrate the importance of a pre-trained feature extractor that is well-adapted to our target space. A pre-trained model with robust feature learning capabilities in the medical or even brain MRI domain can significantly enhance disease monitoring in the medical field.
    
Future work could explore integrating advanced deep learning architectures that improve resolution handling without substantially increasing computational requirements. Further investigations into the training processes might also yield improvements in model stability, particularly for methods like \textbf{EDC} that show potential but are hindered by specific deficiencies. Looking forward, several avenues for enhancing SimpleSliceNet could be explored. Future research could investigate integrating a more sophisticated pre-trained network better tailored to the nuances of medical imaging data. Additionally, advancing fine-tuning techniques and utilizing higher quality datasets could further refine the model’s performance. Another promising area of development is improving the method of aggregating depth-wise information, which could lead to even more precise and effective anomaly detection.

Our evaluations demonstrate that SimpleSliceNet surpasses several state-of-the-art methods in detecting anomalies and outperforms traditional 3D reconstruction-based methods in 3D brain MRI volumes. Notably, it achieves this superior performance with significantly reduced computational time and memory usage. These results highlight SimpleSliceNet’s potential as a robust tool for clinical applications, where efficiency and accuracy are paramount.

\section{Human Subjects Data}
In this work, we used three publicly available medical datasets including BraTS2021, Br35H, and IXI. These datasets do not contain any personally identifiable information about patients, ensuring compliance with privacy standards. We have cited these datasets in the reference section to acknowledge their source and usage appropriately.

%Bibliography
\bibliographystyle{unsrt}  
\bibliography{references}

\begin{thebibliography}{10}

\bibitem{zhang2023destseg}
Xuan Zhang, Shiyu Li, Xi~Li, Ping Huang, Jiulong Shan, and Ting Chen.
\newblock Destseg: Segmentation guided denoising student-teacher for anomaly detection.
\newblock In {\em Proceedings of the IEEE/CVF Conference on Computer Vision and Pattern Recognition}, pages 3914--3923, 2023.

\bibitem{gudovskiy2022cflow}
Denis Gudovskiy, Shun Ishizaka, and Kazuki Kozuka.
\newblock Cflow-ad: Real-time unsupervised anomaly detection with localization via conditional normalizing flows.
\newblock In {\em Proceedings of the IEEE/CVF Winter Conference on Applications of Computer Vision}, pages 98--107, 2022.

\bibitem{guo2023template}
Hewei Guo, Liping Ren, Jingjing Fu, Yuwang Wang, Zhizheng Zhang, Cuiling Lan, Haoqian Wang, and Xinwen Hou.
\newblock Template-guided hierarchical feature restoration for anomaly detection.
\newblock In {\em Proceedings of the IEEE/CVF International Conference on Computer Vision}, pages 6447--6458, 2023.

\bibitem{jewell2022one}
John~Taylor Jewell, Vahid~Reza Khazaie, and Yalda Mohsenzadeh.
\newblock One-class learned encoder-decoder network with adversarial context masking for novelty detection.
\newblock In {\em Proceedings of the IEEE/CVF winter conference on applications of computer vision}, pages 3591--3601, 2022.

\bibitem{bergmann2019mvtec}
Paul Bergmann, Michael Fauser, David Sattlegger, and Carsten Steger.
\newblock Mvtec ad--a comprehensive real-world dataset for unsupervised anomaly detection.
\newblock In {\em Proceedings of the IEEE/CVF conference on computer vision and pattern recognition}, pages 9592--9600, 2019.

\bibitem{liu2023simplenet}
Zhikang Liu, Yiming Zhou, Yuansheng Xu, and Zilei Wang.
\newblock Simplenet: A simple network for image anomaly detection and localization.
\newblock In {\em Proceedings of the IEEE/CVF Conference on Computer Vision and Pattern Recognition}, pages 20402--20411, 2023.

\bibitem{yao2023BGAD}
Xincheng Yao, Ruoqi Li, Jing Zhang, Jun Sun, and Chongyang Zhang.
\newblock Explicit boundary guided semi-push-pull contrastive learning for supervised anomaly detection.
\newblock In {\em Proceedings of the IEEE/CVF Conference on Computer Vision and Pattern Recognition}, pages 24490--24499, 2023.

\bibitem{roth2022towards}
Karsten Roth, Latha Pemula, Joaquin Zepeda, Bernhard Sch{\"o}lkopf, Thomas Brox, and Peter Gehler.
\newblock Towards total recall in industrial anomaly detection.
\newblock In {\em Proceedings of the IEEE/CVF Conference on Computer Vision and Pattern Recognition}, pages 14318--14328, 2022.

\bibitem{salehi2021multiresolution}
Mohammadreza Salehi, Niousha Sadjadi, Soroosh Baselizadeh, Mohammad~H Rohban, and Hamid~R Rabiee.
\newblock Multiresolution knowledge distillation for anomaly detection.
\newblock In {\em Proceedings of the IEEE/CVF conference on computer vision and pattern recognition}, pages 14902--14912, 2021.

\bibitem{guo2023EDC}
Jia Guo, Shuai Lu, Lize Jia, Weihang Zhang, and Huiqi Li.
\newblock Encoder-decoder contrast for unsupervised anomaly detection in medical images.
\newblock {\em IEEE Transactions on Medical Imaging}, 2023.

\bibitem{deng2022reverseST}
Hanqiu Deng and Xingyu Li.
\newblock Anomaly detection via reverse distillation from one-class embedding.
\newblock In {\em Proceedings of the IEEE/CVF Conference on Computer Vision and Pattern Recognition}, pages 9737--9746, 2022.

\bibitem{tien2023revisitingRST}
Tran~Dinh Tien, Anh~Tuan Nguyen, Nguyen~Hoang Tran, Ta~Duc Huy, Soan Duong, Chanh D~Tr Nguyen, and Steven~QH Truong.
\newblock Revisiting reverse distillation for anomaly detection.
\newblock In {\em Proceedings of the IEEE/CVF Conference on Computer Vision and Pattern Recognition}, pages 24511--24520, 2023.

\bibitem{luo2023uad3dbrain}
Guoting Luo, Wei Xie, Ronghui Gao, Tao Zheng, Lei Chen, and Huaiqiang Sun.
\newblock Unsupervised anomaly detection in brain mri: Learning abstract distribution from massive healthy brains.
\newblock {\em Computers in Biology and Medicine}, 154:106610, 2023.

\bibitem{pinaya2022uad3dbrain_transformer}
Walter~HL Pinaya, Petru-Daniel Tudosiu, Robert Gray, Geraint Rees, Parashkev Nachev, Sebastien Ourselin, and M~Jorge Cardoso.
\newblock Unsupervised brain imaging 3d anomaly detection and segmentation with transformers.
\newblock {\em Medical Image Analysis}, 79:102475, 2022.

\bibitem{rudie2021UNETbrain}
Jeffrey~D Rudie, David~A Weiss, John~B Colby, Andreas~M Rauschecker, Benjamin Laguna, Steve Braunstein, Leo~P Sugrue, Christopher~P Hess, and Javier~E Villanueva-Meyer.
\newblock Three-dimensional u-net convolutional neural network for detection and segmentation of intracranial metastases.
\newblock {\em Radiology: Artificial Intelligence}, 3(3):e200204, 2021.

\bibitem{li2021UNETwhitematter}
Hailong Li, Ming Chen, Jinghua Wang, Venkata Sita~Priyanka Illapani, Nehal~A Parikh, and Lili He.
\newblock Automatic segmentation of diffuse white matter abnormality on t2-weighted brain mr images using deep learning in very preterm infants.
\newblock {\em Radiology: Artificial Intelligence}, 3(3):e200166, 2021.

\bibitem{gupta2021slicemodel}
Umang Gupta, Pradeep~K Lam, Greg Ver~Steeg, and Paul~M Thompson.
\newblock Improved brain age estimation with slice-based set networks.
\newblock In {\em 2021 IEEE 18th International Symposium on Biomedical Imaging (ISBI)}, pages 840--844. IEEE, 2021.

\bibitem{gupta2023slicemodelpretrained}
Umang Gupta, Tamoghna Chattopadhyay, Nikhil Dhinagar, Paul~M Thompson, and Greg Ver~Steeg.
\newblock Transferring models trained on natural images to 3d mri via position encoded slice models.
\newblock In {\em 2023 IEEE 20th International Symposium on Biomedical Imaging (ISBI)}, pages 1--5. IEEE, 2023.

\bibitem{br35h}
Ahmed Hamada.
\newblock Br35h: Brain tumor detection 2020, 2020.

\bibitem{IXI}
Ixi dataset.

\bibitem{baid2021rsnabrats}
Ujjwal Baid, Satyam Ghodasara, Suyash Mohan, Michel Bilello, Evan Calabrese, Errol Colak, Keyvan Farahani, Jayashree Kalpathy-Cramer, Felipe~C Kitamura, Sarthak Pati, et~al.
\newblock The rsna-asnr-miccai brats 2021 benchmark on brain tumor segmentation and radiogenomic classification.
\newblock {\em arXiv preprint arXiv:2107.02314}, 2021.

\bibitem{bakas2017advancingbrats}
Spyridon Bakas, Hamed Akbari, Aristeidis Sotiras, Michel Bilello, Martin Rozycki, Justin~S Kirby, John~B Freymann, Keyvan Farahani, and Christos Davatzikos.
\newblock Advancing the cancer genome atlas glioma mri collections with expert segmentation labels and radiomic features.
\newblock {\em Scientific data}, 4(1):1--13, 2017.

\bibitem{menze2014multimodalbrats}
Bjoern~H Menze, Andras Jakab, Stefan Bauer, Jayashree Kalpathy-Cramer, Keyvan Farahani, Justin Kirby, Yuliya Burren, Nicole Porz, Johannes Slotboom, Roland Wiest, et~al.
\newblock The multimodal brain tumor image segmentation benchmark (brats).
\newblock {\em IEEE transactions on medical imaging}, 34(10):1993--2024, 2014.

\bibitem{isensee2019hdbet}
Fabian Isensee, Marianne Schell, Irada Pflueger, Gianluca Brugnara, David Bonekamp, Ulf Neuberger, Antje Wick, Heinz-Peter Schlemmer, Sabine Heiland, Wolfgang Wick, et~al.
\newblock Automated brain extraction of multisequence mri using artificial neural networks.
\newblock {\em Human brain mapping}, 40(17):4952--4964, 2019.

\bibitem{jenkinson2001global}
Mark Jenkinson and Stephen Smith.
\newblock A global optimisation method for robust affine registration of brain images.
\newblock {\em Medical image analysis}, 5(2):143--156, 2001.

\bibitem{jenkinson2002improved}
Mark Jenkinson, Peter Bannister, Michael Brady, and Stephen Smith.
\newblock Improved optimization for the robust and accurate linear registration and motion correction of brain images.
\newblock {\em Neuroimage}, 17(2):825--841, 2002.

\bibitem{deng2009imagenet}
Jia Deng, Wei Dong, Richard Socher, Li-Jia Li, Kai Li, and Li~Fei-Fei.
\newblock Imagenet: A large-scale hierarchical image database.
\newblock In {\em 2009 IEEE conference on computer vision and pattern recognition}, pages 248--255. Ieee, 2009.

\bibitem{dinh2016densityNF}
Laurent Dinh, Jascha Sohl-Dickstein, and Samy Bengio.
\newblock Density estimation using real nvp.
\newblock {\em arXiv preprint arXiv:1605.08803}, 2016.

\bibitem{chopra2005learning}
Sumit Chopra, Raia Hadsell, and Yann LeCun.
\newblock Learning a similarity metric discriminatively, with application to face verification.
\newblock In {\em 2005 IEEE computer society conference on computer vision and pattern recognition (CVPR'05)}, volume~1, pages 539--546. IEEE, 2005.

\bibitem{schroff2015facenet}
Florian Schroff, Dmitry Kalenichenko, and James Philbin.
\newblock Facenet: A unified embedding for face recognition and clustering.
\newblock In {\em Proceedings of the IEEE conference on computer vision and pattern recognition}, pages 815--823, 2015.

\bibitem{kascenas2023role}
Antanas Kascenas, Pedro Sanchez, Patrick Schrempf, Chaoyang Wang, William Clackett, Shadia~S Mikhael, Jeremy~P Voisey, Keith Goatman, Alexander Weir, Nicolas Pugeault, et~al.
\newblock The role of noise in denoising models for anomaly detection in medical images.
\newblock {\em Medical Image Analysis}, 90:102963, 2023.

\bibitem{zhao2022aeflow}
Yuzhong Zhao, Qiaoqiao Ding, and Xiaoqun Zhang.
\newblock Ae-flow: Autoencoders with normalizing flows for medical images anomaly detection.
\newblock In {\em The Eleventh International Conference on Learning Representations}, 2022.

\bibitem{li2021cutpaste}
Chun-Liang Li, Kihyuk Sohn, Jinsung Yoon, and Tomas Pfister.
\newblock Cutpaste: Self-supervised learning for anomaly detection and localization.
\newblock In {\em Proceedings of the IEEE/CVF conference on computer vision and pattern recognition}, pages 9664--9674, 2021.

\bibitem{kascenas2022dae_mri}
Antanas Kascenas, Nicolas Pugeault, and Alison~Q O’Neil.
\newblock Denoising autoencoders for unsupervised anomaly detection in brain mri.
\newblock In {\em International Conference on Medical Imaging with Deep Learning}, pages 653--664. PMLR, 2022.

\bibitem{schlegl2019fanogan}
Thomas Schlegl, Philipp Seeb{\"o}ck, Sebastian~M Waldstein, Georg Langs, and Ursula Schmidt-Erfurth.
\newblock f-anogan: Fast unsupervised anomaly detection with generative adversarial networks.
\newblock {\em Medical image analysis}, 54:30--44, 2019.

\bibitem{van2021ad_mri_chronic}
Kees~M van Hespen, Jaco~JM Zwanenburg, Jan~W Dankbaar, Mirjam~I Geerlings, Jeroen Hendrikse, and Hugo~J Kuijf.
\newblock An anomaly detection approach to identify chronic brain infarcts on mri.
\newblock {\em Scientific Reports}, 11(1):7714, 2021.

\bibitem{pinaya2022fastdiffusion}
Walter~HL Pinaya, Mark~S Graham, Robert Gray, Pedro~F Da~Costa, Petru-Daniel Tudosiu, Paul Wright, Yee~H Mah, Andrew~D MacKinnon, James~T Teo, Rolf Jager, et~al.
\newblock Fast unsupervised brain anomaly detection and segmentation with diffusion models.
\newblock In {\em International Conference on Medical Image Computing and Computer-Assisted Intervention}, pages 705--714. Springer, 2022.

\bibitem{schluter2022NSA}
Hannah~M Schl{\"u}ter, Jeremy Tan, Benjamin Hou, and Bernhard Kainz.
\newblock Natural synthetic anomalies for self-supervised anomaly detection and localization.
\newblock In {\em European Conference on Computer Vision}, pages 474--489. Springer, 2022.

\bibitem{cciccek20163dunet}
{\"O}zg{\"u}n {\c{C}}i{\c{c}}ek, Ahmed Abdulkadir, Soeren~S Lienkamp, Thomas Brox, and Olaf Ronneberger.
\newblock 3d u-net: learning dense volumetric segmentation from sparse annotation.
\newblock In {\em Medical Image Computing and Computer-Assisted Intervention--MICCAI 2016: 19th International Conference, Athens, Greece, October 17-21, 2016, Proceedings, Part II 19}, pages 424--432. Springer, 2016.

\bibitem{lee2017superhumanresunet}
Kisuk Lee, Jonathan Zung, Peter Li, Viren Jain, and H~Sebastian Seung.
\newblock Superhuman accuracy on the snemi3d connectomics challenge.
\newblock {\em arXiv preprint arXiv:1706.00120}, 2017.

\end{thebibliography}

\end{document}